%% file: acl2017.tex
\documentclass[11pt,a4paper]{article}
\usepackage[hyperref]{acl2017}
\usepackage{times}
\usepackage{url}
\usepackage{latexsym}
\usepackage{fixltx2e}
\usepackage{graphicx}
\usepackage{lipsum}

\usepackage{url}
\usepackage{array}
\usepackage{comment}
\usepackage{amssymb}
\usepackage{amsmath}
\usepackage{setspace}

\DeclareMathOperator*{\argmax}{\arg\!\max}
\usepackage{hhline}
\usepackage{multirow}
\usepackage{color}
\usepackage{booktabs}
\usepackage{graphicx}
\usepackage{caption}
\usepackage{lipsum}

\newcommand\blfootnote[1]{%
  \begingroup
  \renewcommand\thefootnote{}\footnote{#1}%
  \addtocounter{footnote}{-1}%
  \endgroup
}
\usepackage[noend]{myalgorithmic}
\usepackage{myalgorithm}

\renewcommand{\algorithmicrequire}{\textbf{Input:}}
\renewcommand{\algorithmicensure}{\textbf{Output:}}
\renewcommand{\algorithmicforall}{\textbf{for each}}
\renewcommand{\algorithmicdo}{}

\def\env@matrix{\hskip -\arraycolsep
  \let\@ifnextchar\new@ifnextchar
  \array{*\c@MaxMatrixCols c}}

\aclfinalcopy 
\def\aclpaperid{***} 

\title{Embedding Words and Senses Together \\ via Joint Knowledge-Enhanced Training}

 \author{Massimiliano Mancini*, Jose Camacho-Collados*, Ignacio Iacobacci \and Roberto Navigli \\
         Department of Computer Science\\
	    Sapienza University of Rome\\
	     {\tt mancini@dis.uniroma1.it} \\
	    {\tt \{collados,iacobacci,navigli\}@di.uniroma1.it}}

\date{}

\begin{document}





\maketitle
\begin{abstract}
  Word embeddings are widely used in Natural Language Processing, mainly due to their success in capturing semantic information from massive corpora. However, their creation process does not allow the different meanings of a word to be automatically separated, as it conflates them into a single vector. We address this issue by proposing a new model which learns word and sense embeddings jointly. Our model exploits large corpora and knowledge from semantic networks in order to produce a unified vector space of word and sense embeddings. We evaluate the main features of our approach both qualitatively and quantitatively in a variety of tasks, highlighting the advantages of the proposed method in comparison to state-of-the-art word- and sense-based models. \blfootnote{Authors marked with an asterisk (*) contributed equally.}
\end{abstract}


\section{Introduction}


Recently, approaches based on neural networks which embed words into low-dimensional vector spaces from text corpora (i.e. word embeddings) have become increasingly popular \cite{Mikolovetal:2013,pennington2014glove}. 
Word embeddings have 
proved to be beneficial in many Natural Language Processing tasks, such as Machine Translation \cite{zou2013bilingual}, syntactic parsing \cite{weiss2015structured}, and Question Answering \cite{bordes2014question}, to name a few. Despite their success in capturing semantic properties of words, these representations are generally hampered by an important limitation: the inability to discriminate among different meanings of the same word.
 
Previous works have addressed this limitation by automatically inducing word senses from monolingual corpora 
\cite{schutze1998automatic,ReisingerMooney:2010,Huangetal:2012,di2013clustering,Neelakantanetal:2014,tian2014probabilistic,LiJurafsky:2015,vu2016k,qiu-tu-yu:2016:EMNLP2016}, or bilingual parallel data 
\cite{guo2014learning,ettinger2016retrofitting,vsuster2016bilingual}. 
However, these approaches learn solely on the basis of statistics extracted from text corpora and do not exploit knowledge from semantic networks. Additionally, their induced senses are neither readily interpretable \cite{panchenko-EtAl:2017:EACLlong} nor easily mappable to lexical resources, which limits their application. Recent approaches have utilized semantic networks to inject knowledge into existing word representations \cite{yu2014improving,faruqui2014retrofitting,goikoetxea2015random,speer-lowryduda:2017:SemEval,Mrksik:17tacl}, 
but without solving the meaning conflation issue. 
In order to obtain a representation for each sense of a word, a number of approaches have leveraged lexical resources to learn sense embeddings as a result of post-processing conventional word embeddings \cite{chenunified:2014,johansson2015embedding,jauhar2015ontologically,RotheSchutze:2015,PilehvarCollier:2016emnlp,camacho2016nasari}. 

Instead, we propose SW2V (\textit{Senses and Words to Vectors}), a neural model that exploits knowledge from both text corpora and semantic networks 
in order to simultaneously learn embeddings for both words and senses. 
Moreover, our model provides three additional key features: (1) both word and sense embeddings are represented in the same vector space, (2) it is flexible, as it can be applied to different predictive models, and (3) it is scalable for very large semantic networks and 
text corpora.

\section{Related work}
\label{Related_work}

Embedding words from large corpora into a low-dimensional vector space has been a popular task since the appearance of the probabilistic feedforward neural network language model \cite{bengio2003neural} and later developments 
such as word2vec \cite{Mikolovetal:2013} and GloVe \cite{pennington2014glove}. 
However, little research has focused on exploiting lexical resources to overcome the inherent ambiguity of word embeddings.


\newcite{iacobacci:2015} overcame this limitation by applying an off-the-shelf disambiguation system (i.e. Babelfy \cite{Moroetal:14tacl}) to a corpus and then using 
word2vec to learn sense embeddings over the pre-disambiguated text. However, in their approach words are replaced by their intended senses, consequently 
producing as output sense representations only. 
The representation of words and senses in the same vector space proves essential for applying these knowledge-based sense embeddings in downstream applications, particularly for their integration into neural architectures \cite{pilehvaracl17}. In the literature, various different methods have attempted to overcome this limitation. 
\newcite{chenunified:2014} proposed a model for obtaining both word and sense representations based on a first training step of conventional word embeddings, a second disambiguation step based on sense definitions, and a final training phase which uses the disambiguated text as input.
Likewise, \newcite{RotheSchutze:2015} aimed at building a shared space of word and sense embeddings based on two steps: a first training step of only word embeddings and a second training step to produce sense and synset embeddings. These two approaches require 
multiple steps of training and make use of a relatively small resource like WordNet, which limits their coverage and applicability. \newcite{camacho2016nasari} increased the coverage of these WordNet-based approaches by exploiting the complementary knowledge of WordNet and Wikipedia along with pre-trained word embeddings. Finally, \newcite{wang2014knowledge} and \newcite{fang2016entity} proposed a model to align vector spaces of words and entities from knowledge bases. However, these approaches are restricted to nominal instances only (i.e. Wikipedia pages or entities).  

In contrast, we propose a model which learns both words and sense embeddings from a single joint training phase, producing a common vector space of words and senses as an emerging feature.


\section{Connecting words and senses in context}
\label{disambiguation}

In order to jointly produce embeddings for words and senses, SW2V needs as input a corpus where words are connected to senses\footnote{In this paper we focus on senses but other items connected to words may be used (e.g. supersenses or images).} 
in each given context. 
One option for obtaining such connections could be to take a sense-annotated corpus as input. 
However, manually annotating large amounts of data is extremely expensive and therefore impractical in normal settings. Obtaining sense-annotated data from current off-the-shelf disambiguation and entity linking systems is possible, but generally suffers from two major problems.
First, supervised systems are hampered by the very same problem of needing large amounts of sense-annotated data. 
Second, the relatively slow speed of current disambiguation systems, such as graph-based approaches \cite{hoffart2012kore,agirre2014random,Moroetal:14tacl}, or word-expert supervised systems \cite{ZhongNg:2010,iacobacci-pilehvar-navigli:2016:P16-1,melamud2016context2vec}, could become an obstacle when applied to large corpora. 


This is the reason why we propose a simple yet effective unsupervised \textit{shallow word-sense connectivity} algorithm, which can be applied to virtually any given semantic network and is linear on the corpus size. 
The main idea of the algorithm is to exploit the connections of a semantic network by associating words with the senses that are most connected within the sentence, according to the underlying network. 



\textbf{Shallow word-sense connectivity algorithm.} Formally, a corpus and a semantic network are taken as input and a set of connected words and senses is produced as output. We define a semantic network as a graph $(S,E)$ where the set $S$ contains synsets (nodes) and $E$ represents a set of semantically connected synset pairs (edges). Algorithm \ref{alg1} describes how to connect words and senses in a given text 
(sentence or paragraph) $T$. First, we gather in a set $S_T$ all candidate synsets of the words (including multiwords up to trigrams) in $T$ (lines \ref{line:beginning} to \ref{line:endfor}). Second, for each candidate synset $s$ we calculate the number of synsets which are connected with $s$ in the semantic network and are included in $S_T$, excluding connections of synsets which only appear as candidates of the same word (lines \ref{line:2ndphase} to \ref{line:numconnections}). Finally, each word is associated with its top candidate synset(s) according to its/their number of connections in context, provided that its/their number of connections exceeds a threshold $\theta=\frac{|S_T|+|T|}{2 \, \delta}$ (lines \ref{line:numconnectionsafter} to \ref{line:end_14}).\footnote{As mentioned above, all unigrams, bigrams and trigrams present in the semantic network are considered. In the case of overlapping instances, the selection of the final instance is performed in this order: mention whose synset is more connected (i.e. $n$ is higher), longer mention and from left to right.} This parameter aims to retain relevant connectivity across senses, as only senses above the threshold will be connected to words in the output corpus. $\theta$ is proportional to the reciprocal of a parameter $\delta$,\footnote{Higher values of $\delta$ lead to higher recall, while lower values of $\delta$ increase precision but lower the recall. We set the value of $\delta$ to $100$, as it was shown to produce a fine balance between precision and recall. This parameter may also be tuned on downstream tasks.} and directly proportional to the average text length and number of candidate synsets within the text. 

\begin{algorithm}[t!]
\caption{Shallow word-sense connectivity} 
\label{alg1}                         
\small
\begin{algorithmic}[1]
  \REQUIRE Semantic network $(S,E)$ and text $T$ represented as a bag of words
  \ENSURE Set of connected words and senses $T^* \subset T \times S$
  
    \STATE {Set of synsets $S_T \gets \emptyset$} \label{line:beginning}  
    
    \FORALL {word $w \in T$} \label{line:mainFor11}       
    
        \STATE {$S_T \gets S_T \cup S_w$ ($S_w$: set of candidate synsets of $w$)}
        
    \ENDFOR \label{line:endfor}

    \STATE{Minimum connections threshold $\theta \gets \frac{|S_T|+|T|}{2 \, \delta}$ } \label{line:threshold} 
    
    \STATE {Output set of connections $T^* \gets \emptyset$}  \label{line:2ndphase} 
    
    \FORALL {$w \in T$}  \label{line:start_14}
    
            \STATE{Relative maximum connections $max=0$}
            \STATE{Set of senses associated with $w$, $C_w \gets \emptyset$}
            
            \FORALL {candidate synset $s \in S_w$} 
                \STATE{Number of edges $n=|s'\in S_T: (s,s')\in E \And \exists w' \in T: w'\ne w \And s' \in S_{w'} |$ } \label{line:numconnections} 
                \IF{$n \geq max  \And  n \geq \theta $} \label{line:numconnectionsafter} 
                    \IF{$n > max$} 
                        \STATE{$C_w \gets \{ (w,s) \}$}
                        \STATE{$max \gets n$}
                        
                    \ELSE
                        \STATE{$C_w \gets C_w \cup \{ (w,s) \}$}
                        
                    \ENDIF
                
                \ENDIF
            \ENDFOR
            
            \STATE{$T^* \gets T^* \cup C_w$}

    \ENDFOR \label{line:end_14}

    \RETURN Output set of connected words and senses $T^*$ \label{line:return1}

\end{algorithmic}
\end{algorithm}

The complexity of the proposed algorithm is $N+(N \times \alpha)$, where $N$ is the number of words of the training corpus and $\alpha$ is the average polysemy degree of a word in the corpus according to the input semantic network. Considering that non-content words are not taken into account (i.e. polysemy degree 0) and that the average polysemy degree of words in current lexical resources (e.g. WordNet or BabelNet) does not exceed a small constant (3) in any language, we can safely assume that the algorithm is linear in the size of the training corpus. Hence, the training time is not significantly increased in comparison to training words only, irrespective of the corpus size. 
This enables a fast training on large amounts of text corpora, in contrast to current unsupervised disambiguation algorithms. Additionally, as we will show in Section \ref{disambiguationtest}, this algorithm does not only speed up significantly the training phase, but also leads to more accurate results.

Note that with our algorithm a word is allowed to have more than one sense associated. In fact, current lexical resources like WordNet \cite{miller1995wordnet} or BabelNet \cite{NavigliPonzetto:12aij} are hampered by the high granularity of their sense inventories \cite{Hovyetal:13}. In Section \ref{clustering} we show how our sense embeddings are particularly suited to deal with this issue.

\section{Joint training of words and senses}
\label{training}


\begin{figure*}[t!]
\begin{center}
    \hspace{-80pt}\includegraphics[scale=0.45]{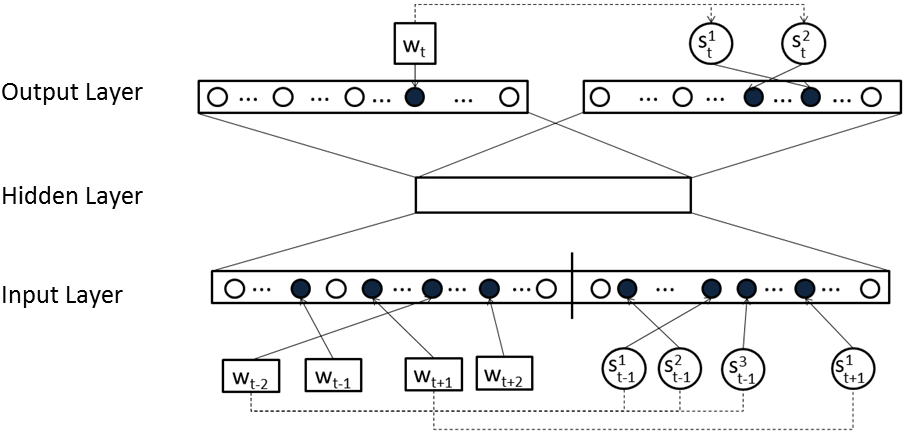}
\end{center}
    \caption{The SW2V architecture on a sample training instance using four context words. Dotted lines represent the 
    virtual link between words and 
    associated senses in context. In this 
    example, the input layer consists of a context of two previous words
    ($w_{t-2}$, $w_{t-1}$) and two subsequent words ($w_{t+1}$, $w_{t+2}$) with respect to the target word $w_t$. Two words ($w_{t-1}$, $w_{t+2}$) do not have senses associated in context, while $w_{t-2}$, $w_{t+1}$ have three senses ($s_{t-1}^1$, $s_{t-1}^2$, $s_{t-1}^3$) and one sense associated ($s_{t+1}^1$) in context, respectively. 
    The output layer consists of the target word $w_t$, which has two senses associated ($s_{t}^1$, $s_{t}^2$) in context.} 
    
    
    \label{fig:input}
\end{figure*}

The goal of our approach is to obtain a shared vector space of words and senses. To this end, our model extends 
conventional word embedding models by integrating explicit knowledge into its architecture. While we will focus on the Continuous Bag Of Words (CBOW) architecture of word2vec \cite{Mikolovetal:2013}, our extension can easily be applied similarly to Skip-Gram
, or to other predictive approaches based on neural networks. The CBOW architecture is based on the feedforward neural network language model \cite{bengio2003neural} and aims at predicting the current word using its surrounding context. The architecture consists of input, hidden and output layers. The input layer has the size of the word vocabulary 
and encodes the context as a combination of one-hot vector representations of surrounding words of a given target word. The output layer has the same size as the input layer and contains a one-hot vector of the target word during the training phase.  


Our model extends the input and output layers of the neural network with word senses\footnote{Our model can also produce a space of words and synset embeddings as output: the only difference is that all synonym senses would be considered to be the same item, i.e. a synset.} by exploiting the intrinsic relationship 
between words and senses. 
The leading principle is that, since a word is the surface form of an underlying sense, updating the embedding of the word should produce a consequent update to the embedding representing that particular sense, and vice-versa. 
As a consequence of the algorithm described in the previous section, each word in the corpus may be connected with zero, one or more senses. We refer to the set of senses connected to a given word within the specific context as its \textit{associated senses}. 

%

Formally, we define a training instance as a sequence of words $W=w_{t-n},...,w_t,...,w_{t+n}$ (being $w_t$ the target word) 
and $S=S_{t-n},...,S_t,....,S_{t+n}$, where $S_i=s_i^1,...,s_i^{k_i}$ is the sequence of all associated senses 
in context of $w_i \in W$. 
Note that $S_i$ might be empty if the word $w_i$ does not have any associated sense.
In our model each target word takes as context both its surrounding words and all the senses associated with them. 
In contrast to the original CBOW architecture, where the training criterion is to correctly classify $w_t$, our approach aims to predict
the word $w_t$ and its set $S_t$ of associated senses.  
This is equivalent to minimizing the following loss function:
\begin{equation*}
E=-\log(p(w_{t} | W^t  , S^t )) -\sum_{s\in{S_t}}\log(p(s | W^t , S^t  )) \\
\end{equation*}
\noindent where $W^t=w_{t-n},...,w_{t-1}, w_{t+1},...,w_{t+n}$ and $S^t=S_{t-n},...,S_{t-1},S_{t+1},...,S_{t+n}$. Figure \ref{fig:input} shows the organization of the input and the output layers on a sample training instance. In what follows we present a set of variants of the model on the output and the input layers.

    
    


\subsection{Output layer alternatives}
\label{outputalternatives}

\begin{description}
\item [Both words and senses.] This is the default case explained above. If a word has one or more associated senses
, these senses are also used as target on a separate output layer. 
\item [Only words.] In this case we exclude senses as target. There is a single output layer with the size of the word vocabulary as in the original CBOW model.
\item [Only senses.] 
In contrast, this alternative excludes words, using only senses as target. In this case, if a word does not have any associated sense, it is not used as 
target instance.
\end{description}

\subsection{Input layer alternatives}
\label{inputalternatives}

\begin{description}

\item [Both words and senses.] Words and their associated senses are included in the input layer and contribute to the hidden state. Both words and senses are updated as a consequence of the backpropagation algorithm.
\item [Only words.] In this alternative only the surrounding words contribute to the hidden state, i.e. the target word/sense (depending on the alternative of the output layer) is predicted only from word features. 
The update of an input word is propagated to the embeddings of its associated senses, if any. In other words, despite not being included in the input layer, senses still receive the same gradient of the associated input word, through a virtual connection. 
This configuration, coupled with the only-words output layer configuration, corresponds exactly to the default CBOW architecture of word2vec with the only addition of the update step for senses.
\item [Only senses.] Words are excluded from the input layer and the target is predicted only from the senses associated with the surrounding words. 
The weights of the words are updated through the updates of the associated senses, in contrast to the only-words alternative.



\end{description}

\begin{table*}[t]

\begin{center}

    \scalebox{0.85}{\begin{tabular}{| l | c | c|c | c|c || c|c| c|c || c|c |c|c| }
     \cline{3-14}
    \multicolumn{1}{c}{} & \multicolumn{1}{c|}{} & \multicolumn{12}{c|}{\textbf{Output}}  \\
        \cline{3-14}
        
       \multicolumn{1}{c}{} & \multicolumn{1}{c|}{} & \multicolumn{4}{c||}{\textbf{Words}} & \multicolumn{4}{c||}{\textbf{Senses}} & \multicolumn{4}{c|}{\textbf{Both}}  \\
       
          \cline{3-14}
        
       \multicolumn{1}{c}{} & \multicolumn{1}{c|}{} &  \multicolumn{2}{c|}{\textbf{WS-Sim}} & \multicolumn{2}{c||}{\textbf{RG-65}} &
        \multicolumn{2}{c|}{\textbf{WS-Sim}} & \multicolumn{2}{c||}{\textbf{RG-65}} 
        &  \multicolumn{2}{c|}{\textbf{WS-Sim}} & \multicolumn{2}{c|}{\textbf{RG-65}} \\

        \cline{3-14}
        \multicolumn{1}{c}{} & \multicolumn{1}{c|}{} & \multicolumn{1}{c|}{\textbf{$r$}}  &  \multicolumn{1}{c|}{\textbf{$\rho$}} & \multicolumn{1}{c|}{\textbf{$r$}}  &  \multicolumn{1}{c||}{\textbf{$\rho$}} & \multicolumn{1}{c|}{\textbf{$r$}}  &  \multicolumn{1}{c|}{\textbf{$\rho$}}  & \multicolumn{1}{c|}{\textbf{$r$}}  &  \multicolumn{1}{c||}{\textbf{$\rho$}} & \multicolumn{1}{c|}{\textbf{$r$}}  &  \multicolumn{1}{c|}{\textbf{$\rho$}} & \multicolumn{1}{c|}{\textbf{$r$}}  &  \multicolumn{1}{c|}{\textbf{$\rho$}}  \\ 
        
         
         
        \hline
         \cline{2-14}
          \multirow{3}{*}{\rotatebox[origin=c]{90}{\bf Input}} &  \textbf{Words} &  0.49 &  0.48 &  0.65 &  0.66 &  0.56 &  0.56 & 0.67 & 0.67 & 0.54 & 0.53 & 0.66 & 0.65\\
         \cline{2-14}
         & \textbf{Senses}   &  0.69 &  0.69 & 0.70 &  0.71 &   0.69 &  0.70 & 0.70 & \textbf{0.74} & \textbf{0.72} & \textbf{0.71} & \textbf{0.71} & \textbf{0.74}\\
        \cline{2-14}
        & \textbf{Both}  &  0.60 &  0.65 & 0.67 &  0.70 &  0.62 &  0.65 & 0.66 & 0.67 & 0.65 & \textbf{0.71} & 0.68 & 0.70\\
        
        \hline
    \end{tabular}
}

\end{center}
    \caption{Pearson ($r$) and Spearman ($\rho$) correlation performance of the nine configurations of SW2V }
    \label{tab:simanalysis}
\end{table*}

\section{Analysis of Model Components}
\label{analysis}

In this section we analyze the different components of SW2V, including the nine model configurations (Section \ref{parameters}) and the algorithm which generates the connections between words and senses in context (Section \ref{disambiguationtest}).
In what follows we describe the common analysis setting:

\begin{itemize}

\item \textbf{Training model and hyperparameters.} For evaluation purposes, we use the CBOW model of word2vec with standard hyperparameters: the dimensionality of the vectors is set to 300 and the window size to 8, and hierarchical softmax is used for normalization. These hyperparameter values are set across all experiments. 

\item \textbf{Corpus and semantic network.} We use a 300M-words corpus from the UMBC project \cite{han2013umbc}, which contains English paragraphs extracted from the web.\footnote{\url{http://ebiquity.umbc.edu/blogger/2013/05/01/umbc-webbase-corpus-of-3b-english-words/}} 
As semantic network we use BabelNet 3.0\footnote{\url{http://babelnet.org}}, a large multilingual semantic network with over 350 million semantic connections, integrating resources such as Wikipedia and WordNet. We chose BabelNet owing to its wide coverage of named entities and lexicographic knowledge. 


\item \textbf{Benchmark.} Word similarity has been one of the most popular benchmarks for \textit{in-vitro} evaluation of vector space models \cite{pennington2014glove,levy2015improving}. For the analysis we use two word similarity datasets: the similarity portion \cite[WS-Sim]{Agirreetal:09}
of the WordSim-353 dataset \cite{Levetal:2002} 
and RG-65 \cite{RG65:1965}.
In order to compute the similarity of two words using our sense embeddings, we apply the standard closest senses strategy \cite{Resnik:95,BudanitskyHirst:06,camachocolladosetal:2015b}, using cosine similarity ($\cos$) as comparison measure between senses: 

\end{itemize}
%
\begin{equation}
sim(w_1,w_2)= \max_{s \in {S_{w_1}}, s' \in S_{w_2}} \cos(\vec{s}_1,\vec{s}_2)
\end{equation}
%
\setlength{\leftskip}{0.9cm}
\noindent where $S_{w_i}$ represents the set of all candidate senses of $w_i$ and $\vec{s}_i$ refers to the sense vector representation of the sense $s_i$. 
\newline

\setlength{\leftskip}{0pt}

\subsection{Model configurations}
\label{parameters}

In this section we analyze the different configurations of our model in respect of the input and the output layer on a word similarity experiment. Recall from Section \ref{training} that our model could have words, senses or both in either the input and output layers. Table \ref{tab:simanalysis} shows the results of all nine configurations on the WS-Sim and RG-65 datasets.

As shown in Table \ref{tab:simanalysis}, the best configuration according to both Spearman and Pearson correlation measures is the configuration which has only senses in the input layer and both words and senses in the output layer.\footnote{In this analysis we used the word similarity task for optimizing the sense embeddings, without caring about the performance of word embeddings or their interconnectivity. Therefore, this configuration may not be optimal for word embeddings and may be further tuned on specific applications. More information about different configurations in the documentation of the source code.} In fact, taking only senses as input seems to be consistently the best alternative for the input layer. Our hunch is that the knowledge learned from both the co-occurrence information and the semantic network is more balanced with this input setting. For instance, in the case of including both words and senses in the input layer, the co-occurrence information learned by the network would be duplicated for both words and senses.

\subsection{Disambiguation / Shallow word-sense connectivity algorithm }
\label{disambiguationtest}

In this section we evaluate the impact of our \textit{shallow word-sense connectivity algorithm} (Section \ref{disambiguation}) by testing our model directly taking a pre-disambiguated text as input. In this case the network exploits the connections between each word and its disambiguated sense in context. 
For this comparison we used Babelfy\footnote{\url{http://babelfy.org}} \cite{Moroetal:14tacl}, a state-of-the-art graph-based disambiguation and entity linking system based on BabelNet. 
We compare to both the default Babelfy system which uses the \textit{Most Common Sense} (MCS) heuristic as a back-off strategy and, following \cite{iacobacci:2015}, we also include a version in which only instances above the Babelfy default confidence threshold are disambiguated (i.e. the MCS back-off strategy is disabled). We will refer to this latter version as Babelfy* and report the best configuration of each strategy according to our analysis. 

Table \ref{tab:disanalysis} shows the results of our model using the three different strategies on RG-65 and WS-Sim. Our shallow word-sense connectivity algorithm achieves the best overall results. We believe that these results are due to the semantic connectivity ensured by our algorithm and to the possibility of associating words with more than one sense, which seems beneficial for training, making it more robust to possible disambiguation errors and to the sense granularity issue \cite{erk2013measuring}. The results are especially significant considering that our algorithm took a tenth of the time needed by Babelfy to 
process the corpus.


\begin{table}[t]
 
\begin{center}
\scalebox{0.95}{
    

    \begin{tabular}{| l | c|c || c|c |}
        \cline{2-5}
         \multicolumn{1}{c|}{} & \multicolumn{2}{c||}{\textbf{WS-Sim}} & \multicolumn{2}{c|}{\textbf{RG-65}}  \\
        \cline{2-5}  
        \multicolumn{1}{c|}{} & \multicolumn{1}{c|}{\textbf{$r$}}  &  \multicolumn{1}{c||}{\textbf{$\rho$}} & \multicolumn{1}{c|}{\textbf{$r$}}  &  \multicolumn{1}{c|}{\textbf{$\rho$}}   \\ 
         
        
        \hline
          \textit{Shallow} &  \bf 0.72 &  \bf 0.71 &  \bf 0.71  &  \bf 0.74   \\
         \cline{1-5}
          Babelfy~ &  0.65 &  0.63 &  0.69 &  0.70 \\
         \cline{1-5}
          Babelfy*   &  0.63 &  0.61  &  0.65  &  0.64 \\
        \cline{1-5}

        \hline
    \end{tabular}
}

\end{center}
    \caption{Pearson ($r$) and Spearman ($\rho$) correlation performance of SW2V integrating our \textit{shallow} word-sense connectivity algorithm (default), Babelfy, or Babelfy*.}
    \label{tab:disanalysis}
\end{table}

\section{Evaluation}
\label{sec:evaluation}

\begin{table*}

\begin{center}
{
\begin{tabular}{|l|l|c| c| c|c| c|}
\cline{4-7}
\multicolumn{3}{c|}{}	&  \multicolumn{2}{c|}{\bf SimLex-999} &\multicolumn{2}{c|}{\bf MEN}  \\
\cline{2-7}
\multicolumn{1}{c|}{} & \multicolumn{1}{c|}{\textbf{System}} & \multicolumn{1}{c|}{\textbf{Corpus}}	&  \multicolumn{1}{c|}{\textbf{$r$}}  &  \multicolumn{1}{c|}{\textbf{$\rho$}} 	&  \multicolumn{1}{c|}{\textbf{$r$}}  &  \multicolumn{1}{c|}{\textbf{$\rho$}}  \\
\hline
 \multirow{8}{*}{\textbf{Senses}}
& SW2V\textsubscript{BN}	& UMBC &  \bf  0.49  &  \bf 0.47  	&  0.75  &   0.75       \\
\cline{2-7}
& SW2V\textsubscript{WN}	& UMBC &  0.46  &  0.45  	&  \bf  0.76  &    \bf 0.76      \\
\cline{2-7}
& AutoExtend & UMBC 	&    0.47    &  0.45  & 0.74 &  0.75  \\
\cline{2-7}
& AutoExtend & Google-News 	&        0.46      &  0.46   & 0.68 &  0.70  \\
\cline{2-7}
\cline{2-7}
& SW2V\textsubscript{BN} 	& Wikipedia &    0.47  &   0.43 	&    0.71 &   0.73      \\
\cline{2-7}
& SW2V\textsubscript{WN} 	& Wikipedia &    0.47  &   0.43 	&    0.71 &   0.72      \\
\cline{2-7}
& SensEmbed  & Wikipedia    	&         0.43       &  0.39  & 0.65 &   0.70             \\
\cline{2-7}
& Chen et al. (2014) & Wikipedia    	&         0.46        &  0.43  & 0.62 & 0.62               \\
 \hline
 \hline
\multirow{6}{*}{\textbf{Words}}
 & Word2vec & UMBC 	&        0.39    &     0.39  &  0.75  &   0.75      \\
\cline{2-7}
& Retrofitting\textsubscript{BN} & UMBC	&     0.47    &  0.46  &  0.75  &     \bf 0.76  \\
 \cline{2-7}
 & Retrofitting\textsubscript{WN} & UMBC	&     0.47    &  0.46  &  \bf 0.76  &     \bf 0.76  \\
 \cline{2-7}
  & Word2vec & Wikipedia   	&        0.39    &     0.38  &  0.71  &  0.72      \\
\cline{2-7}
& Retrofitting\textsubscript{BN} & Wikipedia  	&    0.35    &  0.32 &  0.66   &     0.66   \\
\cline{2-7}
& Retrofitting\textsubscript{WN} & Wikipedia	&    0.47    &  0.44  &  0.73  &    0.73  \\

 \hline
\end{tabular}
}
\end{center}
\caption{\label{tab:simlex} Pearson ($r$) and Spearman ($\rho$) correlation performance on the SimLex-999 and MEN word similarity datasets.}
\end{table*}

We perform a qualitative and quantitative evaluation of 
important features of SW2V in three different tasks. First, in order to compare our model against standard word-based approaches, we evaluate our system in the word similarity task (Section \ref{wordsimilarity}). Second, we measure the quality of our sense embeddings in a sense-specific application: sense clustering (Section \ref{clustering}). Finally, we evaluate the coherence of our unified vector space by measuring the interconnectivity of word and sense embeddings (Section \ref{predominant}). 

\textbf{Experimental setting.} Throughout all the experiments we use the same standard hyperparameters mentioned in Section \ref{analysis} for both the original word2vec implementation and our 
proposed model SW2V. For SW2V we use the same optimal configuration according to the analysis of the previous section (only senses as input, and both words and senses as output) for all tasks. As training corpus we take the full 3B-words UMBC webbase corpus and the Wikipedia (Wikipedia dump of November 2014), used by three of the comparison systems. We use BabelNet 3.0 (SW2V\textsubscript{BN}) and WordNet 3.0 (SW2V\textsubscript{WN}) as semantic networks. 

\textbf{Comparison systems.} We compare with the publicly available pre-trained sense embeddings of four state-of-the-art models: \newcite{chenunified:2014}\footnote{
\url{http://pan.baidu.com/s/1eQcPK8i}} and AutoExtend\footnote{
We used the AutoExtend code (\url{http://cistern.cis.lmu.de/~sascha/AutoExtend/}) to obtain sense vectors using W2V embeddings trained on UMBC (GoogleNews corpus used in their pre-trained models is not publicly available). We also tried the code to include BabelNet as lexical resource, but it was not easily scalable (BabelNet is two orders of magnitude larger than WordNet).} \cite{RotheSchutze:2015} based on WordNet, 
and SensEmbed\footnote{
\url{http://lcl.uniroma1.it/sensembed/}} \cite{iacobacci:2015} and NASARI\footnote{\url{http://lcl.uniroma1.it/nasari/}}  
\cite{camacho2016nasari} based on BabelNet.

\subsection{Word Similarity}
\label{wordsimilarity}

In this section we evaluate our sense representations on the standard SimLex-999 \cite{hill2015simlex} and MEN \cite{bruni2014multimodal} word similarity datasets\footnote{To enable a fair comparison we did not perform experiments on the small datasets used in Section \ref{analysis} for validation.}. SimLex and MEN contain 999 and 3000 word pairs, respectively, which constitute, to our knowledge, the two largest similarity datasets 
comprising a balanced set of noun, verb and adjective instances. 
As explained in Section \ref{analysis}, we use the closest sense strategy for the word similarity measurement of our model and all 
sense-based comparison systems. As regards the word embedding models, words are directly compared by using cosine similarity. 
We also include a \textit{retrofitted} version of the original word2vec word vectors \cite[Retrofitting\footnote{\url{https://github.com/mfaruqui/retrofitting}}]{faruqui2014retrofitting} using WordNet (Retrofitting\textsubscript{WN}) and BabelNet (Retrofitting\textsubscript{BN}) as lexical resources.

Table \ref{tab:simlex} shows the results of SW2V and all comparison models in SimLex and MEN. SW2V consistently outperforms all sense-based comparison systems using the same corpus, and clearly performs better than the original word2vec trained on the same corpus. Retrofitting decreases the performance of the original word2vec on the Wikipedia corpus using BabelNet as lexical resource, but significantly improves the original word vectors on the UMBC corpus, obtaining comparable results to our approach. However, while our approach provides a shared space of words and senses, Retrofitting still conflates different meanings of a word into the same vector. 

Additionally, we noticed that most of the score divergences between our system and the gold standard scores in SimLex-999 were produced on antonym pairs, which are over-represented in this dataset: 38 word pairs hold a clear antonymy relation (e.g. \textit{encourage-discourage} or \textit{long-short}), while 41 additional pairs hold some degree of antonymy (e.g. \textit{new-ancient} or \textit{man-woman}).\footnote{Two annotators decided the degree of antonymy between word pairs: \textit{clear antonyms}, \textit{weak antonyms} or \textit{neither}.} 
In contrast to the consistently low gold similarity scores given to antonym pairs, our system varies its similarity scores depending on the specific nature of the pair\footnote{For instance, the pairs \textit{sunset-sunrise} and \textit{day-night} are given, respectively, 1.88 and 2.47 gold scores in the 0-10 scale, while our model gives them a higher similarity score. In fact, both pairs appear as coordinate synsets in WordNet.}. 
Recent works have managed to obtain significant improvements by tweaking usual word embedding approaches into providing low similarity scores for antonym pairs \cite{marcobaroni2multitask,schwartz2015symmetric,nguyen:2016:antsyn, Mrksik:17tacl}, 
but this is outside the scope of this paper.


\subsection{Sense Clustering}
\label{clustering}

Current lexical resources tend to suffer from the high granularity of their sense inventories \cite{Palmeretal:07}. In fact, a meaningful clustering of their senses may lead to improvements on downstream tasks \cite{Hovyetal:13,flekovasupersense,pilehvaracl17}. 
In this section we evaluate our synset representations on the Wikipedia sense clustering task. 
For a fair comparison with respect to the BabelNet-based comparison systems that use the Wikipedia corpus for training, in this experiment we report the results of our model trained on the Wikipedia corpus and using BabelNet as lexical resource only.
For the evaluation we consider the two Wikipedia sense clustering datasets (500-pair and SemEval) created by \newcite{Dandalaetal:2013}. In these datasets sense clustering is viewed as a binary classification task in which, given a pair of Wikipedia pages, the system has to decide whether to cluster them into a single instance or not. To this end, we use our synset embeddings and cluster Wikipedia pages\footnote{
Since Wikipedia is 
a resource included in BabelNet, our synset representations are expandable to Wikipedia pages. } together if their similarity 
exceeds a threshold $\gamma$. In order to set the optimal value of $\gamma$, we follow Dandala et al. (2013) and use the first 500-pairs sense clustering dataset for tuning. We set the threshold $\gamma$ to 0.35, which is the value leading to the highest F-Measure among all values from 0 to 1 with a 0.05 step size on the 500-pair dataset. Likewise, we set a threshold for NASARI (0.7) and SensEmbed (0.3) comparison systems.

\begin{table}
\begin{center}
\centering
\scalebox{1.0}{

\begin{tabular}{|l |c|c|}
\cline{2-3}
\multicolumn{1}{l|}{} &  \multicolumn{1}{c|}   {\bf Accuracy}   & \multicolumn{1}{c|} {\bf F-Measure}  
\\
    \hline 
    SW2V	&	\bf 87.8	& \bf 63.9 	\\ 
    \hline 

    SensEmbed 	&	 82.7	& 40.3 	\\
    \hline 
    \textsc{NASARI}	&	 87.0	& 62.5 	\\
    \hline 
    Multi-SVM  &	85.5	& - \\
    \hline 
    Mono-SVM  &	83.5	& - \\
    \hline 
     
Baseline    	&	17.5	& 29.8 \\
       \hline
\end{tabular} 
}
\end{center}
\caption{\label{tab:clustering} Accuracy and F-Measure percentages of different systems on the SemEval Wikipedia sense clustering dataset.}
\end{table}

Finally, we evaluate our approach on the SemEval sense clustering test set. This test set consists of 925 pairs which were obtained from a set of highly ambiguous words gathered from past SemEval tasks. For comparison, we also include the supervised approach of \newcite{Dandalaetal:2013} based on a multi-feature Support Vector Machine classifier trained on an automatically-labeled dataset of the English Wikipedia (Mono-SVM) and Wikipedia in four different languages (Multi-SVM). As naive baseline we include the system which would cluster all given pairs.  

Table \ref{tab:clustering} shows the F-Measure and accuracy results 
on the SemEval sense clustering dataset. SW2V outperforms all comparison systems according to both measures, including the sense representations of NASARI and SensEmbed using the same setup and the same underlying lexical resource. 
This confirms the capability of our system to accurately capture the semantics of word senses on this sense-specific task.



\subsection{Word and sense interconnectivity}
\label{predominant}

In the previous experiments we evaluated the effectiveness of the sense embeddings. In contrast, this experiment aims at testing the interconnectivity between word and sense embeddings in the vector space. As explained in Section \ref{Related_work}, there have been previous approaches building a shared space of word and sense embeddings
, but to date little research has focused on testing the semantic coherence of the vector space.
To this end, we evaluate our model on a Word Sense Disambiguation (WSD) task, using our shared vector space of words and senses to obtain a \textit{Most Common Sense} (MCS) baseline. The insight behind this experiment is that a semantically coherent shared space of words and senses should be able to build a relatively strong baseline for the task, as the MCS of a given word should be closer to the word vector than any other sense. 
The MCS baseline is generally integrated into the pipeline of state-of-the-art WSD and Entity Linking systems as a back-off strategy \cite{navigli:09,jin2009estimating,ZhongNg:2010,Moroetal:14tacl,raganato-camachocollados-navigli:2017:EACLlong} and is used in various NLP applications \cite{bennett2016lexsemtm}. Therefore, a system which automatically identifies the MCS of words from non-annotated text may be quite valuable, especially for resource-poor languages or large knowledge resources for which obtaining sense-annotated corpora is extremely expensive. Moreover, even in a resource like WordNet for which sense-annotated data is available \cite[SemCor]{Milleretal:93}, 61\% of its polysemous lemmas have no sense annotations \cite{bennett2016lexsemtm}.  

Given an input word $w$, we compute the cosine similarity between $w$ and all its candidate senses, picking the sense leading to the highest similarity:
\begin{equation}
MCS(w)=\argmax_{s \in S_w} {\cos(\vec{w},\vec{s})}
\end{equation}
\noindent where $\cos(\vec{w},\vec{s})$ refers to the cosine similarity between the  
embeddings of $w$ and $s$. In order to assess the reliability of SW2V against previous models using WordNet as sense inventory, we test our model on the all-words SemEval-2007 (task 17) \cite{Pradhanetal:07a} and SemEval-2013 (task 12) \cite{Naviglietal:13} WSD datasets. 
Note that our model using BabelNet as semantic network has a far larger coverage than just WordNet and may additionally be used for Wikification \cite{mihalcea:07b} and Entity Linking tasks. Since the versions of WordNet vary across datasets and comparison systems, we decided to evaluate the systems on the portion of the datasets covered by all comparison systems\footnote{We were unable to obtain the word embeddings of \newcite{chenunified:2014} for comparison even after contacting the authors.} (less than 10\% of instances were removed from each dataset). 

\begin{table}
\begin{center}
{
\scalebox{0.9}{ \begin{tabular}{|l| c| c|}
\cline{2-3}
\multicolumn{1}{c|}{} 	&	\multicolumn{1}{c|}{\bf SemEval-07}	& \multicolumn{1}{c|}{\bf  SemEval-13}	\\ 


\hline
SW2V & \bf 39.9 & \bf 54.0	\\
\hline
AutoExtend   &  17.6 &   31.0    \\
\hline
Baseline   	& 24.8	& 34.9    \\
\hline

\end{tabular}
}
}
\end{center}
\caption{\label{tab:mfs} F-Measure percentage of different MCS strategies on the SemEval-2007 and SemEval-2013 WSD datasets.
}
\end{table}

Table \ref{tab:mfs} shows the results of our system and AutoExtend 
on the SemEval-2007 and SemEval-2013 WSD datasets. SW2V provides the best MCS results in both datasets. In general, AutoExtend does not accurately capture the predominant sense of a word and performs worse than a baseline that selects the intended sense randomly from the set of all possible senses of the target word. 

In fact, AutoExtend tends to create clusters which include a word and all its possible senses. As an example, Table \ref{tab:closest} shows the closest word and sense\footnote{Following \newcite{navigli:09}, \textit{word}$_n^p$ is the $n^{th}$ sense of \textit{word} with part of speech $p$ (using WordNet 3.0).} embeddings of our SW2V model and AutoExtend to the \textit{military} and \textit{fish} senses of, respectively, \textit{company} and \textit{school}. 
AutoExtend creates clusters with all the senses of \textit{company} and \textit{school} and their related instances, even if they belong to different domains 
(e.g., firm$_n^2$ or business$_n^1$ clearly concern the \textit{business} sense of \textit{company)}. Instead, SW2V creates a semantic cluster of word and sense embeddings which are semantically close to the corresponding \textit{company$_n^2$} and \textit{school$_n^7$} senses.

\begin{table}
\begin{center}
{
\scalebox{0.82}{
\begin{tabular}{c|c||c|c}
\multicolumn{2}{c}{ \textit{company$_n^2$ (military unit)}}	& \multicolumn{2}{c}{  \textit{school$_n^7$ (group of fish)}}	 \\ [1ex]
	\multicolumn{1}{c|}{\underline{\bf AutoExtend}}	& \multicolumn{1}{c||}{\underline{\bf SW2V}} & \multicolumn{1}{c|}{\underline{\bf AutoExtend}}	& \multicolumn{1}{c}{\underline{\bf SW2V}}\\ 


company$_n^9$  & battalion$_n^1$ & school & schools$_n^7$ \\
company & battalion	& school$_n^4$ & sharks$_n^1$\\
company$_n^8$  &   regiment$_n^1$ & school$_n^6$ & sharks \\
company$_n^6$ &  detachment$_n^4$ & school$_v^1$ & shoals$_n^3$	 	\\
company$_n^7$    	& platoon$_n^1$ & school$_n^3$	& fish$_n^1$  \\
company$_v^1$	& brigade$_n^1$	& elementary  & dolphins$_n^1$ \\
firm  	&  regiment	 & schools & pods$_n^3$ \\
business$_n^1$  	& corps$_n^1$ & elementary$_a^3$ & eels \\
firm$_n^2$  	& brigade & school$_n^5$ & dolphins	  \\
company$_n^1$ 	& platoon & elementary$_a^1$ & whales$_n^2$ 	  \\

\end{tabular}
}
}

\end{center}
\caption{\label{tab:closest} Ten closest word and sense embeddings to the senses \textit{company$_n^2$} (military unit) and \textit{school$_n^7$} (group of fish).
}
\end{table}

\section{Conclusion and Future Work}

In this paper we proposed SW2V (\textit{Senses and Words to Vectors})
, a neural model which learns vector representations for words and senses in a joint training phase by exploiting both text corpora and knowledge from semantic networks. Data (including the preprocessed corpora and pre-trained embeddings used in the evaluation) and source code to apply our extension of the word2vec architecture to learn word and sense embeddings from any preprocessed corpus are freely available at \url{http://lcl.uniroma1.it/sw2v}. 
Unlike previous sense-based models which require 
post-processing steps and use WordNet as sense inventory, 
our model achieves a semantically coherent vector space of both words and senses as an emerging feature of a single training phase and is easily scalable to larger semantic networks like BabelNet. 
Finally, we showed, both quantitatively and qualitatively, some of the advantages of using our approach as against previous state-of-the-art word- and sense-based models in various tasks, and highlighted interesting semantic properties of the resulting unified vector space of word and sense embeddings. 

As future work we plan to integrate a WSD and Entity Linking system for applying our model on downstream NLP applications, along the lines of \newcite{pilehvaracl17}. 
We are also planning to apply 
our model to languages other than English and to study its potential on multilingual and cross-lingual applications.


\input{acknowledgements}

\bibliography{acl2017}
\bibliographystyle{acl_natbib}

\end{document}

%% file: acknowledgements.tex

\section*{Acknowledgments}

\vspace{1ex}
\noindent
\begin{minipage}{0.1\linewidth}
  \raisebox{-0.3\height}{\includegraphics[trim =32mm 55mm 30mm 5mm, clip, scale=0.23]{figs/erc.ai}}
\end{minipage}
\hspace{0.01\linewidth}
\begin{minipage}{0.72\linewidth}
  The authors gratefully acknowledge the support of the ERC Consolidator Grant MOUSSE No. 726487.
\end{minipage}
\hspace{0.01\linewidth}
\begin{minipage}{0.05\linewidth}
  \vspace{0.05cm}
\raisebox{-0.25\height}{\includegraphics[trim =0mm 5mm 5mm 2mm,clip,scale=0.085]{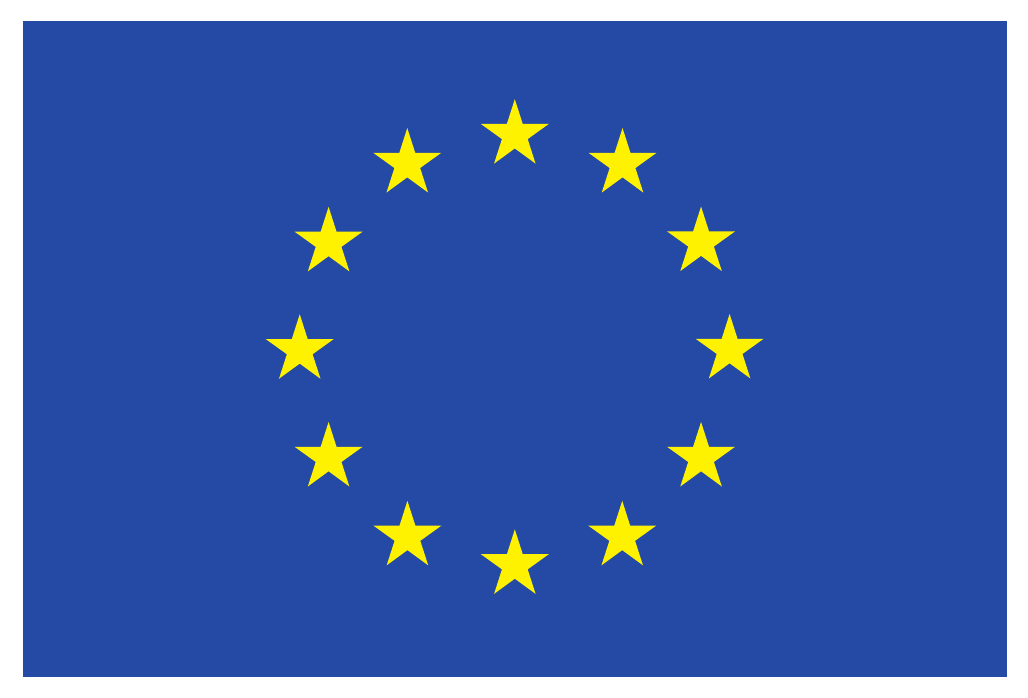}}
\end{minipage}
\vspace{2ex}

Jose Camacho-Collados is supported by a Google Doctoral Fellowship in Natural Language Processing. We would also like to thank Jim McManus for his comments on the manuscript.